# Prediction of Activated Sludge Settling Characteristics from Microscopy Images with Deep Convolutional Neural Networks and Transfer Learning

**Authors:**

Sina Borzooei[*] (corresponding author)
sina.borzooei@ivl.se
Orcid: 0000-0002-0694-3064
IVL Swedish Environmental Research Institute, P.O. Box 210 60, Stockholm SE-100 31, Sweden/ BIOMATH, Department of Data Analysis and Mathematical Modelling, Faculty of Bioscience Engineering, Ghent University, Coupure links 653, 9000 Gent, Belgium

Leonardo Scabini
scabini@ifsc.usp.br
São Carlos Institute of Physics, University of São Paulo, 13560-970, São Carlos - SP, Brazil/ KERMIT, Department of Data Analysis and Mathematical Modelling, Ghent University, Coupure links 653, 9000 Gent, Belgium

Gisele Miranda
gmirand@kth.se
Division of Computational Science and Technology, KTH Royal Institute of Technology, 10044 Stockholm, Sweden

Saba Daneshgar
saba.daneshgar@ugent.be
BIOMATH, Department of Data Analysis and Mathematical Modelling, Faculty of Bioscience Engineering, Ghent University, Coupure links 653, 9000 Gent, Belgium/ CAPTURE, Centre for Advanced Process Technology for Urban Resource Recovery, Frieda Saeysstraat 1, 9000 Gent, Belgium

Lukas Deblieck
lukas.deblieck@pantarein.be
Pantarein Water, Egide Walschaertsstraat 22L, 2800 Mechelen, Belgium

Piet De Langhe
piet.delanghe@pantarein.be
Pantarein Water, Egide Walschaertsstraat 22L, 2800 Mechelen, Belgium

Odemir Bruno
bruno@ifsc.usp.br
São Carlos Institute of Physics, University of São Paulo, 13560-970, São Carlos - SP, Brazil

Bernard De Baets
Bernard.DeBaets@ugent.be
ORCID: 0000-0002-3876-620X
KERMIT, Department of Data Analysis and Mathematical Modelling, Ghent University, Coupure links 653, 9000 Gent, Belgium

Ingmar Nopens
Ingmar.nopens@ugent.be
BIOMATH, Department of Data Analysis and Mathematical Modelling, Faculty of Bioscience Engineering, Ghent University, Coupure links 653, 9000 Gent, Belgium/ CAPTURE, Centre for Advanced Process Technology for Urban Resource Recovery, Frieda Saeysstraat 1, 9000 Gent, Belgium

Elena Torfs
elena.torfs@ugent.be
Université Laval, 1065 avenue de la Médecine, Québec G1V 0A6, QC, Canada/ BIOMATH, Department of Data Analysis and Mathematical Modelling, Ghent University, Coupure links 653, 9000 Gent, Belgium

**Conflict of interest: The authors declare that they have no conflict of interest**

# Abstract

Microbial communities play a key role in biological wastewater treatment processes. Activated sludge settling characteristics, for example, are affected by microbial community composition, varying by changes in operating conditions and influent characteristics of wastewater treatment plants (WWTPs). Timely assessment and prediction of changes in microbial composition leading to settling problems, such as filamentous bulking (FB), can prevent operational challenges, reduction of treatment efficiency, and adverse environmental impacts. This study presents an innovative computer vision-based approach to assess activated sludge-settling characteristics based on the morphological properties of flocs and filaments in microscopy images. Implementing the transfer learning of deep convolutional neural network (CNN) models, this approach aims to overcome the limitations of existing quantitative image analysis techniques. The offline microscopy image dataset was collected over two years, with weekly sampling at a full-scale industrial WWTP in Belgium. Multiple data augmentation techniques were employed to enhance the generalizability of the CNN models. Various CNN architectures, including Inception v3, ResNet18, ResNet152, ConvNeXt-nano, and ConvNeXt-S, were tested to evaluate their performance in predicting sludge settling characteristics. The sludge volume index (SVI) was used as the final prediction variable, but the method can easily be adjusted to predict any other settling metric of choice. The results showed that the suggested CNN-based approach provides less labour-intensive, objective, and consistent assessments while transfer learning notably minimises the training phase, resulting in a generalizable system that can be employed in real-time applications. The model's capability to signal early indicators of FB events makes it a valuable tool for effective monitoring and preemptive management of WWTPs facing activated sludge settling problems.

# Nomenclature

| | |
|---|---|
| ASP | Activated Sludge Process |
| AI | Artificial Intelligence |
| ANN | Artificial Neural Network |
| APHA | American Public Health Association |
| BNRAS | Biological Nutrient Removal Activated Sludge |
| CNN | Convolutional Neural Network |
| CM | Computer Vision |
| COD | Total Chemical Oxygen Demand |
| DL | Deep Learning |
| DO | Dissolved Oxygen concentration |
| FB | Filamentous Bulking |
| FLOPs | FLoating point OPerations |
| IS | Image Segmentation |
| MAE | Mean Absolute Error |
| MAPE | Mean Absolute Percentage Error |
| MBR | Membrane BioReactor |
| ML | Machine Learning |
| MLSS | Mixed Liquor Suspended Solids |
| MSE | Mean Squared Error |
| MTD | Mean Tweedie Deviance |
| PID | Proportional-Integral-Derivative |
| QIA | Quantitative Image Analysis |
| SBR | Sequencing Batch Reactor |
| SRT | Solids Retention Time |
| SVI | Sludge Volume Index |
| TFS | Trained From Scratch |
| TL | Transfer Learning |
| TN | Total Nitrogen |
| TP | Total Phosphorous |
| TSS | Total Suspended Solid |
| ViT | Vision Transformer |
| WAS | Wasted Activated Sludge |
| WWTP | Wastewater Treatment Plant |

## 1. Introduction

The Activated Sludge Process (ASP) is the most commonly used suspended-growth biological treatment technology for municipal and industrial wastewater treatment. The ASP is a two-stage process, including biochemical and physical solid separation stages in the settling and clarification units. Available microorganisms in activated sludge break down the complex substances into simple, stable substances and consume carbon, ammonium and phosphate from the wastewater. Therefore, the core of the ASP is the maintenance of mixed microbial cultures in sludge. Flocs formed by the floc-forming species in ASP can be separated from treated wastewater by gravity forces as one of the most energy-efficient and economic solid-liquid separation technologies (Henze et al., 2008). This is essential to maintain sufficient biomass retention time and sludge concentration in the biochemical stage. Good sludge settleability and compatibility are crucial to ensure efficient solid-liquid separation in secondary clarifiers at modern Biological Nutrient Removal Activated Sludge (BNRAS) systems. A balance between the different types of microorganisms is necessary to guarantee efficient nutrient removal, good sludge-settling characteristics, and effective solid removal. Typical sludge settling issues in BNRAS systems are pinpoint flocs formation, filamentous bulking and foaming, dispersed growth, and viscous bulking (Amaral et al., 2013). Filamentous bacteria such as Sphaerotilus natans and Candidatus Microthrix parvicella (M. parvicella) are the backbone of floc formation and a structure to floc-forming bacteria. When filamentous bacteria overgrow due to various physicochemical factors and/or changes in process conditions, it can result in open flocs and/or inter-floc bridging (Burger et al., 2017). This phenomenon, called Filamentous Bulking (FB), is the most commonly occurring solids settling problem, which can deteriorate flocs settling characteristics, pose operational challenges and failure, and increase treatment costs (Lu et al., 2023). FB is reported as the number one cause of effluent noncompliance in WWTPs with opposing economic and environmental impacts (Richard et al., 2003). It is known that over 50% of WWTPs worldwide reported operational issues due to FB (Martins et al., 2004). Despite extensive research on FB, it appears to be a persistent issue in WWTPs. This can be because a thorough and precise understanding of FB is hampered by the complex relationships between the various microorganism species, their diversity, and the lack of clarity surrounding operational conditions inducing the phenomena (Sam et al., 2022). In wastewater treatment practice, sludge bulking can be characterized by a high Sludge Volume Index (SVI) (e.g. SVI > 150 ml $g^{-1}$) (Martins et al., 2004). However, the critical SVI value above which bulking can occur is highly linked to the local operation, practice and design of secondary clarifiers (Henze et al., 2008). This makes FB an empirical phenomenon without an accurate scientific and quantitative index to distinguish bulking from non-bulking sludge.

Recently, various optical microscopy techniques have been widely used in BNRAS systems to investigate the morphology and composition of activated sludge flocs and identify different microbial species and arrangements of microorganisms (Costa et al., 2022). Identifying and characterizing filamentous bacteria provides the most crucial information for understanding FB. This identification is primarily based on the morphological characterization of filamentous bacteria and their responses to a few microscopic staining tests. Initially, human-performed determination of the floc size and filament length using a counting chamber was implemented on the offline microscope observations to distinguish the flocs and filaments based on various size ranges (Cenens et al., 2002). As these manual methods are time-demanding, subjectively interpretable and prone to human error, automated Quantitative Image Analysis (QIA) tools have been employed to monitor ASPs (Costa et al., 2013). Several studies demonstrated successful applications of QIA techniques for determining filamentous bacteria contents, analyzing the aggregate morphology and floc structures to assess the AS operating parameters using phase-contrast, bright-field and fluorescence microscopy images or a combination thereof (Banadda et al., 2003; Burger et al., 2017; Campbell et al., 2019; Costa et al., 2022; Da Motta et al., 2001; Dias et al., 2016; Koivuranta et al., 2015; Mesquita et al., 2016, 2010; Oliveira et al., 2018; Silva et al., 2022). However, QIA applications usually require laborious and time-consuming image pre/processing steps such as image enhancement, object segmentation, debris and background detection and elimination. In the reviewed QIA applications for investigating FB phenomena, conventional image segmentation (IS) methods such as thresholding, histogram-based bundling and region growing were implemented to differentiate between background, bulky objects, filaments and flocs. These semi-automatic techniques can be highly influenced by the subjectivity and bias of practitioners' decisions regarding the selection of algorithms and the pre-defined threshold values. Among various implemented IS methods, none can be regarded as the most appropriate technique for all types of sludge images. The method designed for a specific final application and the particular image type might not necessarily apply to other cases. Researchers who lack computer science expertise encounter a challenge in correctly parametrizing and using these methods.

Additionally, the final goal of the majority of reviewed QIA applications was to identify the filamentous bacteria using different morphological indicators and ultimately investigate linear/nonlinear correlations among them, standard wastewater treatment operating parameters and/or settleability indices such as SVI (Banadda et al., 2003; Boztoprak et al., 2016; Mesquita et al., 2009). This requires a well-trained practitioner to decide the most relevant morphological parameters to be extracted for each application. For instance, Amaral and Ferreira (2005) reported a group of 36 morphological descriptors, such as the aspect ratio, roundness, reduced radius of gyration, form factor, etc., to be checked to establish relationships between macroscopic and microscopic properties of the BNRAS system. Each descriptor encapsulates different aspects of the microbial community. The further challenge concerns considering combinations of these descriptors in multivariate analyses to solve the complexities of floc formation, FB, and their subsequent impacts on system performance. Identifying a single descriptor that fully encapsulates all factors affecting settling characteristics presents a significant challenge, if not impossible. Hence, considering a microscopy image as an integral entity is a viable alternative which requires advanced computational technologies.

The morphology of filamentous bacteria in BNRAS systems can vary due to environmental and operational changes, complicating their detection tasks. To cope with this complexity and given the current transitional trend of the water industry towards digitalization due to the fourth industrial revolution (Industry 4.0) (Bai et al., 2020), there is an urgent need for advanced automated solutions such as the deep learning (DL) paradigm. DL is a branch of machine learning (ML) based on Artificial Neural Networks (ANNs) designed as a multiple-layer processing architecture that can be used to obtain high-level features from input data, including but not limited to images, sounds, videos, and text (Goodfellow et al., 2016). DL-based techniques have become the de-facto methodology in Computer Vision (CV) applications. By combining preprocessing, segmentation, feature extraction and prediction into a unifed framework, DL allows less human interventions regarding fine-tuning CV models (Chai et al., 2021). The Convolutional Neural Network (CNN) is a class of DL methods commonly used for processing grid-structured data, such as images. CNNs were designed to learn feature representations from low to high-level patterns through convolution operations with filter-defined sizes. Such architecture requires extensive training on large datasets to adaptively learn spatial hierarchies of features through backpropagation (Ghabri et al., 2023). A common approach is to employ transfer learning (TL) by using a model already trained in a specific dataset and then fine-tuning it for different tasks and inpute data (Chan et al., 2023). Successful applications of TL were demonstrated for the classification of cervical cancer microscopy images (Nguyen et al., 2018), classification of microbeads in urban wastewater (Yurtsever and Yurtsever, 2019), concrete crack detection (Dung, 2019) and tea disease detection (Ramdan et al., 2020). In the context of ASPs, the only available application of TL of CNNs was reported by Satoh et al. (2021), where the authors proposed two separate CNNs formulated for supervised classification of binary labelled images. The models successfully detected aggregated and dispersed flocs and the abundance of filamentous bacteria.

Inspired by the success of CNNs in CV, this paper presents a CNN-based approach for predicting activated sludge settling characteristics based on the morphological properties of flocs and filaments found in offline microscopy images. This is achieved by applying pre-trained CNN models on extensive image datasets (ImageNet) and fine-tuning them using microscopy images from a real-scale industrial WWTP. Multiple data augmentation techniques are employed to generate supplementary training data to enhance the generalizability of the CNN models. Various CNN architectures, including Inception v3, ResNet18, ResNet152, ConvNeXt-nano, and ConvNeXt-S, are examined, and their performance in predicting sludge settling parameters, such as the SVI, is compared.

## 2. Materials and methods

### 2.1 Data collection and preparation

The dataset used in this study was collected from a WWTP servicing a maltery industry in Antwerp, Belgium. Under the operation of Pantarein Water Bvba, the WWTP employs an intricate, multi-stage treatment process designed to handle the wastewater produced during the malting process efficiently. The treatment process includes a cylindrical sequencing batch reactor (SBR) with a BNRAS system, followed by separation in a membrane bioreactor (MBR). The SBR unit with a 6 m depth and a volume of 5200 $m^3$ treats an average of 2300 $m^3$ of wastewater influent daily. An on-off controller for blower activation and a proportional-integral-derivative (PID) controller regulating blower frequency are used in the SBR unit's aerating system to control the dissolved oxygen (DO) concentration. The waste-activated sludge (WAS) is removed from the SBR at an average flow rate of 170 $m^3$/day. Weekly samples from the SBR unit were collected over an extended period, from November 1, 2020, to July 31, 2022. During this time, the sludge retention time (SRT) was kept, on average, 4.7 days with a standard deviation of 5.3 days, indicating different operational states. Physical, chemical, and microscopic analyses were performed on the samples. The biomass's settling capacity was determined for each sample using a cylindrical column, with sludge height variation observations conducted for 30 minutes within a 1L settling cylinder. Total Suspended Solids (TSS) were quantified by weight determination post-filtering a defined volume of the AS mixture to separate the liquid and solid phases. This filtration process utilized pre-weighed filter papers (chm, 125 mm, Spain), a Büchner funnel, and a single-stage vacuum pump. After filtration, the residual solid matter and filter were dried in an oven set at 103-105 °C for a duration exceeding 1 hour. The dried filter and sludge were subsequently weighed, and the filter paper's weight was subtracted to determine the AS's dry weight in the original specific volume. TSS measurements were then utilized to calculate the SVI based on the APHA (American Public Health Association) protocol (Beutler et al., 2014). In addition to the data already mentioned, wastewater quality parameters from the influent of WWTP, influent and effluent of SBR, along with the operational parameters of the SBR, were collected and monitored during the study period, as presented in Table 1.

**Table 1**. Overview of wastewater characteristics and SBR operational parameters

| Parameters | Location of sampling/Measuring | Unit | Average ± SD |
|---|---|---|---|
| Flowrate ($Q_{in}$) | Influent | $m^3/d$ | 2300±600 |
| Total Chemical Oxygen Demand ($COD_{in}$) | Influent | mg/L | 2950±500 |
| Total Nitrogen ($TN_{in}$) | Influent | mg/L | 74±20 |
| Total Phosphorous ($TP_{in}$) | Influent | mg/L | 22±7 |
| Total Suspended Solids ($TSS_{in}$) | Influent | mg/L | 350±140 |
| Mixed Liquor Suspended Solids (MLSS) | SBR unit | g/L | 11.2±1.5 |
| pH | SBR unit | - | 6.5±0.5 |
| Total Cycle Time | SBR unit | min | 200±30 |

To obtain microscopy images, a Biohit Proline pipette was used to deposit a 12 μL sample onto a microscope slide covered with an 18 mm×18 mm coverslip. Phase contrast microscopy was employed to capture images of the flocs and filaments on the slide, utilizing an Olympus BX43 microscope with a 4× objective magnification. An average of five images were taken per sample to comprehensively cover the coverslip's surface. The number and distribution of flocs on the slide determined the precise number of images acquired per sample. All images were subsequently resized to a resolution of 512 x 384 pixels, a reduction to one-fourth of the original resolution, to enhance computational efficiency. Pixel normalization was achieved through zero-centering and unity variance (z-score) by the ImageNet mean and variance parameters.

**2.2 Applied CNN models**

CNNs are ANN topologies frequently used in CV problems. They extract the most relevant characteristics by performing a series of operations on incoming data using convolutional and pooling layers. A defined-size filter is convolved with the input data to generate a new matrix in the convolution operation, which is the building block of a CNN. Each convolutional layer is made up of a collection of filters that are used to produce an activation map for each input. The pooling layers are emplyed to reduce the spatial size of the convolution output. Maximum or average pooling are commonly used to compress the input and decrease noise, therefore highlighting the most important features. After the final convolution, the resulting matrix is downscaled into a one-dimensional vector, which is then fed to a feed-forward ANN. The last layer, be it for classification or regression, is selected based on the desired task (Li et al., 2021). The training of a CNN consists of a series of iterations using backpropagation.

Over the last 20 years, different CNN architectures have been proposed. One example is the Inception model, initially called GoogLeNet. It was introduced in an attempt to overcome the computational costs of earlier deep models, mainly designed to stack large convolution operations. Instead, the inception architecture is organized in modules, characterized by parallel convolutional layers in which the signal is independently processed to be finally concatenated. In this way, the network becomes *wider* rather than *deeper*. This is achieved using different filter sizes, characterizing multiscale feature learning, and 1x1 convolutions applied after the max pooling layer for dimensionality reduction. Through concatenating the inception modules, the network reaches greater depths with reduced computational costs compared to previous architectures like AlexNet and VGG. InceptionV3 (Szegedy et al., 2016) is employed in this work, incorporating advancements over the original architecture, such as regularization with batch-normalized auxiliary classifiers and label-smoothing.

The ResNet architecture (He et al., 2016) was proposed to overcome problems related to the "vanishing gradient" problem since the performance of previous models did not scale with the increase in number of network layers. The name comes from residual connections, also known as skip connections, and has the advantage of improving the signal flow and reducing the occurrence of exploding gradients, usually caused by exponential decay or the increase of the signal's magnitude, since the gradient multiplication becomes enormous over time. High gradients result in abrupt changes in the model's weights during training. As a result, the model is unable to learn, leading the training to stagnate. Introducing residual connections reduces this behaviour considerably and allows training deep models more efficiently by reducing the network into fewer layers. In this work, two variations of the ResNet architecture were considered, a small and a bigger one, considering their computational cost: ResNet18 and ResNet152.

After the ResNets were proposed, most of the CNN architectures remained similar, composed of the same design concepts previously described until the proposal of the Vision Transformer (ViT) (Dosovitskiy et al., 2021). ConvNeXts, for instance, are the result of updates on the architecture of a typical ResNet by including ViT principles in the convolutional modules since the arrival of ViTs brought new insights into the CNN study (Liu et al., 2022). ConvNeXts architecture includes patching over input, the ResNeXt structure, an inverted bottleneck, higher filter size, GELu activation functions, enhanced layer normalization, and separate downsampling layers. We investigated two ConvNeXt versions in our experiments, ConvNeXt-nano and ConvNeXt-S, with different computational costs.

One crucial characteristic of CNNs when implementing them in real-world applications is their computational complexity, also known as computational cost or budget. Several metrics can be used to quantify the cost of ANNs. One of the most adopted ones is the number of Floating Point Operations (FLOPs), which measures the computational complexity of a model and estimates its processing time. Another metric consists of counting the parameters (weights), which estimates a model's memory consumption since weight values must be stored. A higher number of FLOPs and parameters may result in a more robust and more accurate model, but it requires increasingly more data and computational resources to train and run. Details of each CNN architecture employed in this study are given in Table 1.

Table 2. Comparison of CNN architectures: depth, computational complexity, and parameters

| Model | Depth | GFLOPs (Giga FLOPs) | Parameters (million) |
|---|---|---|---|
| Inception V3 | 48 | 2.85 | 21.79 |
| ResNet18 | 18 | 1.82 | 11.18 |
| ResNet152 | 152 | 11.58 | 58.15 |
| ConvNeXt-nano | 42 | 2.46 | 14.95 |
| ConvNext-S | 108 | 8.70 | 49.46 |

From Table 2, it can be deduced that regarding GFLOPs, giga FLOPs per second, ResNet152 is the most computationally complex model, followed by ConvNeXt-S. ResNet18 is the least complex model among the five listed. Regarding the number of parameters, ResNet152 is again the largest model, with 58.15 million parameters. ConvNeXt-S has the second-highest number of parameters, followed by Inception v3. ResNet18 has the smallest number of parameters, with 11.18 million.

## 2.3. Data augmentation

Data augmentation is commonly employed to enhance the performance of CNNs during training, mainly when working with limited sample sizes. The technique generates additional samples by applying label-preserving image transformations. This process enriches the dataset and improves the model's generalizability while reducing the risk of overfitting (Shorten and Khoshgoftaar, 2019). For instance, random rotations and translations of training samples can enhance object recognition tasks. However, the selection of augmentation methods should be done carefully to avoid introducing noise. For example, random rotations in medical imaging may lead to anatomically implausible configurations. This study employed data augmentation techniques that preserve critical attributes of floc and filaments morphological characteristics for CNN analysis. Suitable modifications included random flips along both axes, rotations within a range of -180º to 180º, brightness adjustments varying from -20% to +20%, and random erasure of 2% to 20% of the image area. The augmentations were applied randomly during each training iteration, allowing the CNN model to handle variations in floc and filament position, brightness, and potential overlap. Figure 1 displays 12 instances of augmentations implemented on a specific image.

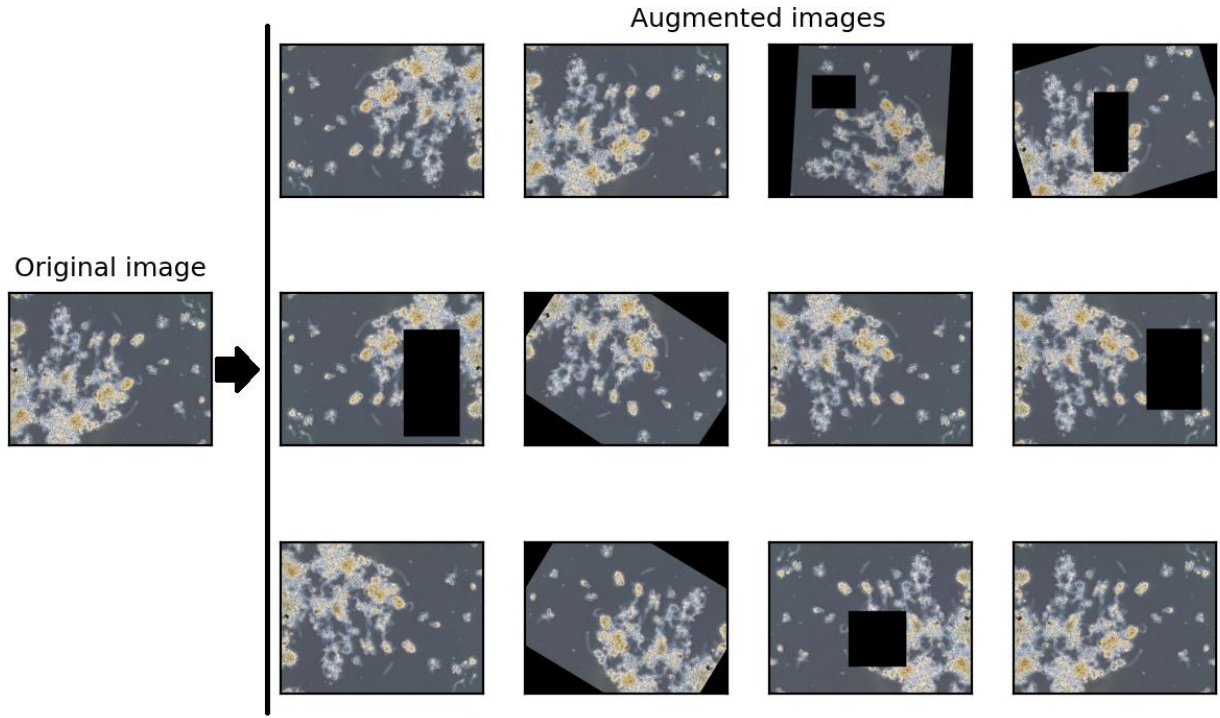

**Fig. 1. Examples of data augmentations applied to a microscopy image for improving CNN training Dataset variability and robustness.**

## 2.4 Model training and transfer learning

In this study, we employed three CNN architectures and their variations to predict SVI values based on microscopy images directly. To this end, the last layers of CNNs were replaced by a linear regression layer. Each CNN architecture was built from scratch or fine-tuned with pre-training on ImageNet-1k (Deng et al., 2009), a well-known computer vision dataset of 1.2 million images categorized into 1000 classes. However, for fine-tuning ConvNeXt-S architecture, a more extensive dataset was required; hence, ImageNet-21k, a dataset with over 14 million images categorized into more than 21,000 classes, was used. Following the cutting-edge ANN training methods proposed by Liu et al. (2022), the final layer of the pre-trained models was initialised with random weights and then trained on the target dataset. The backpropagation approach reduced the Mean Squared Error (MSE) by adopting the AdamW optimizer for loss function minimization, with an initial learning rate of 0.0001 that was adaptable to the depth of the model. Models were fine-tuned using a batch size of 32 across 30 epochs, and the learning rate was annealed throughout the epochs using cosine annealing decay. Employing stochastic depth, layers were randomly omitted during training to reduce parameter numbers and improve model generalization. As a regularisation procedure, weight decay was implemented by adding a penalty term to the loss function to lower the weights and thereby mitigate overfitting. Experiments were conducted using Python (version 3.7) with PyTorch and timm libraries on an Ubuntu server equipped with two Nvidia GeForce RTX 2080ti graphics cards (11 GB VRAM each) and an AMD Ryzen Threadripper 3990x processor with 94GB RAM.

## 2.5. Validation and model evaluation

The MSE was adopted as the loss function used during the training for regression of the CNNs models evaluated in this work. The MSE was also used to assess the model's generalization performance on the test dataset. The models were trained for 30 epochs when using TL approach and 95 epochs when training from scratch. The lowest MSE on the test set defined the final checkpoint. To mitigate overfitting and obtain a robust and reliable performance evaluation, the 10-fold cross-validation method was used. The original dataset was divided into ten subsets. The model was then interatively trained on nine subsets and evaluated on the remaining one, ensuring that each sample in the dataset was used as a test sample at least once. The mean performance was calculated across all folds. The best model is assessed on the corresponding test set during each cross-validation iteration, considering regression metrics other than MSE. The Mean Absolute Error (MAE), Mean Tweedie Deviance (MTD), coefficient of determination ($R^2$), and Mean Absolute Percentage Error (MAPE) were used to evaluate the generalization performance of the trained models quantitatively. In our analysis, $x_i$ represents the ground truth SVI for input image $I$ and $y_i$ represents the model's corresponding prediction output. The MAE is calculated by taking the average absolute difference between the predicted and actual values for each test sample in every cross-validation fold.

$$\text{MAE} = \frac{1}{n}\sum_{i=1}^{n}|x_i - y_i|$$

MAPE is computed as the average of the absolute percentage errors, and $R^2$ measures the fraction of variance in the ground truth values explained by the predicted values:

$$\text{MAPE} = \frac{1}{n}\sum_{i=1}^{n}\frac{|x_i - y_i|}{x_i}$$

$$R^2 = 1 - \frac{\sum_{i=1}^{n}(x_i - y_i)^2}{\sum_{i=1}^{n}\left(x_i - \frac{1}{n}\sum_{j=1}^{n}x_j\right)^2}$$

MTD measures the deviation between the predicted and ground truth values for a Poisson distribution:

$$\text{MTD} = \frac{1}{n}\sum_{i=1}^{n}2\left(x_i \log\left(\frac{x_i}{y_i}\right) + y_i - x_i\right)$$

The average and standard deviation are computed across the ten cross-validation iterations for each metric.

## 3. Results and discussion

### 3.1 CNN comparison and selection of the best model

The training loss (MSE) trajectories of different CNN architectures analyzed during the training iterations are presented in Fig. 2. For all models, as the number of epochs increase, the training loss decreases, indicating no big oscillations, convergence and stabilization of weights, complying with an expected behaviour when training CNNs (Liu et al., 2022). As a general observation, the performance gap between the curves at the beginning of the training phase indicates that TL starts from a lower loss for all CNN models, leading to a better final performance (lower loss at the end of training). Additionally, it can be seen that training the CNNs from scratch requires more epochs (100) compared to the TL approach, which also yields faster learning (30 epochs). When comparing the performance of ResNet architectures, it can be observed that ResNet152 (both TL and non-TL) performed slightly better than ResNet18, suggesting the deeper model fits the training data better. Although the ResNet18 model began with a higher loss when using TL, the situation reversed at the end of training. About ConvNeXt models, both variants start from a very high training loss followed by a sharp drop over the initial epochs, indicating faster learning or adjustment of weights for the training data. Better fitting performance of the TL approach is even more highlighted for the ConvNeXt models. When used with TL, the ConvNeXt-n showed better fitting and lower loss at the end of training than ConvNeXt-s. In this stage, it was concluded that TL is necessary for achieving better SVI fitting to the training data.

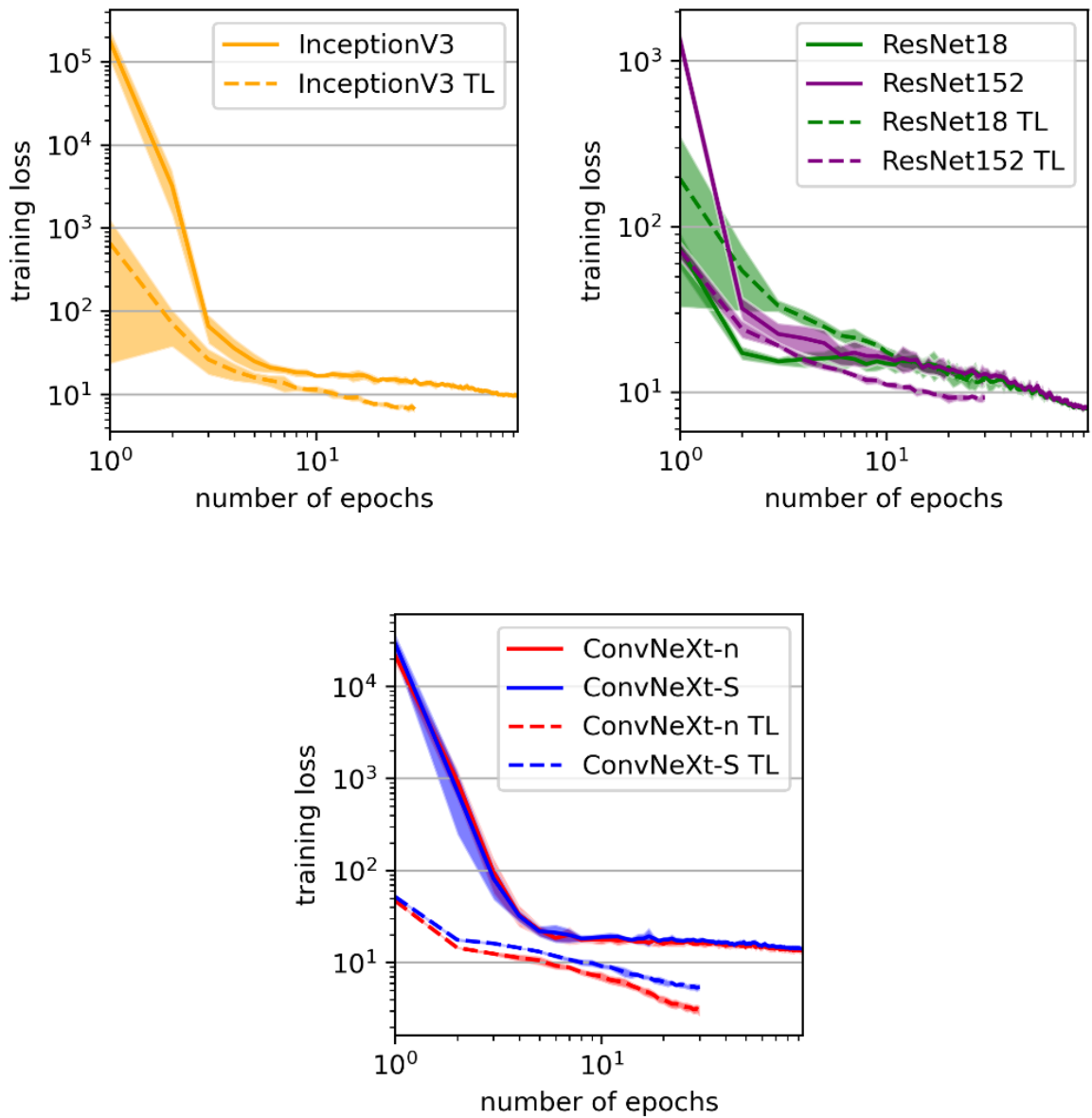

**Fig. 2. Comparative training behaviour of the three CNN architectures when trained from scratch or with TL from ImageNet**

In the next step, we analyzed the performance of the selected CNN architectures that were fine-tuned with only TL on the validation datasets (Fig. 3). Both ConvNeXt-nano and ConvNeXt-S present the lowest validation losses while maintaining stable performance. This result aligns with other works in the CNN literature (Han et al., 2022) since ConvNeXts, developed in 2022, incorporates more advanced techniques (see Section 2.2) than previous models, such as Inception and ResNets.

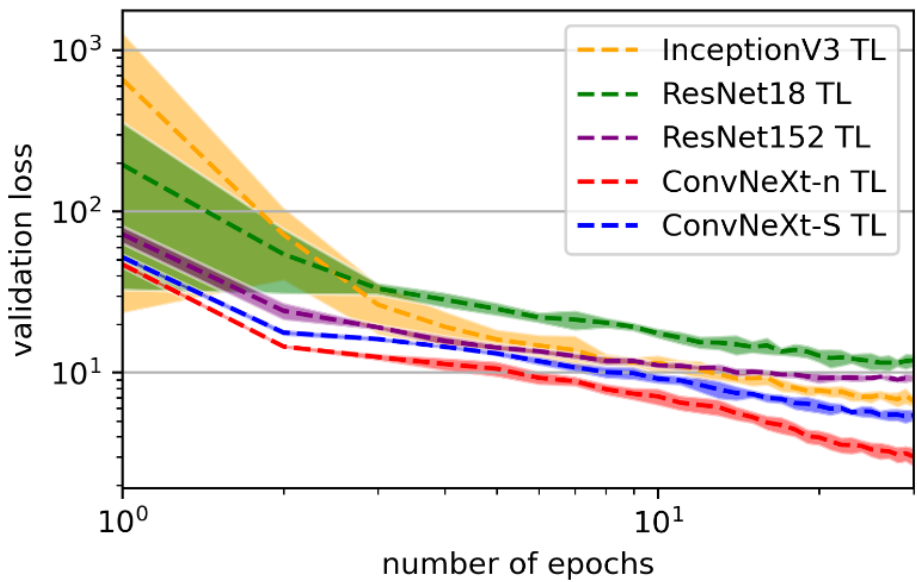

**Fig. 3 Performance comparison of different CNNs, with TL, over the validation splits**

Additionally, bigger CNN variants like ResNet152 and ConvNeXt-S were found to be more challenging to train due to the occurrence of overfitting (evidenced by the gap between training and validation losses), achieving lower generalization performance than their smaller versions, ResNet18 and ConvNeXt-nano. This indicates that added model complexity does not necessarily improve performance. Increasing the quantity of the training data may improve this issue in future. The CNN models fine-tuned with TL and TFS using the validation data set, were compared using four error measures on the test set: MAE, MTD, $R^2$, and MAPE (Fig. 4).

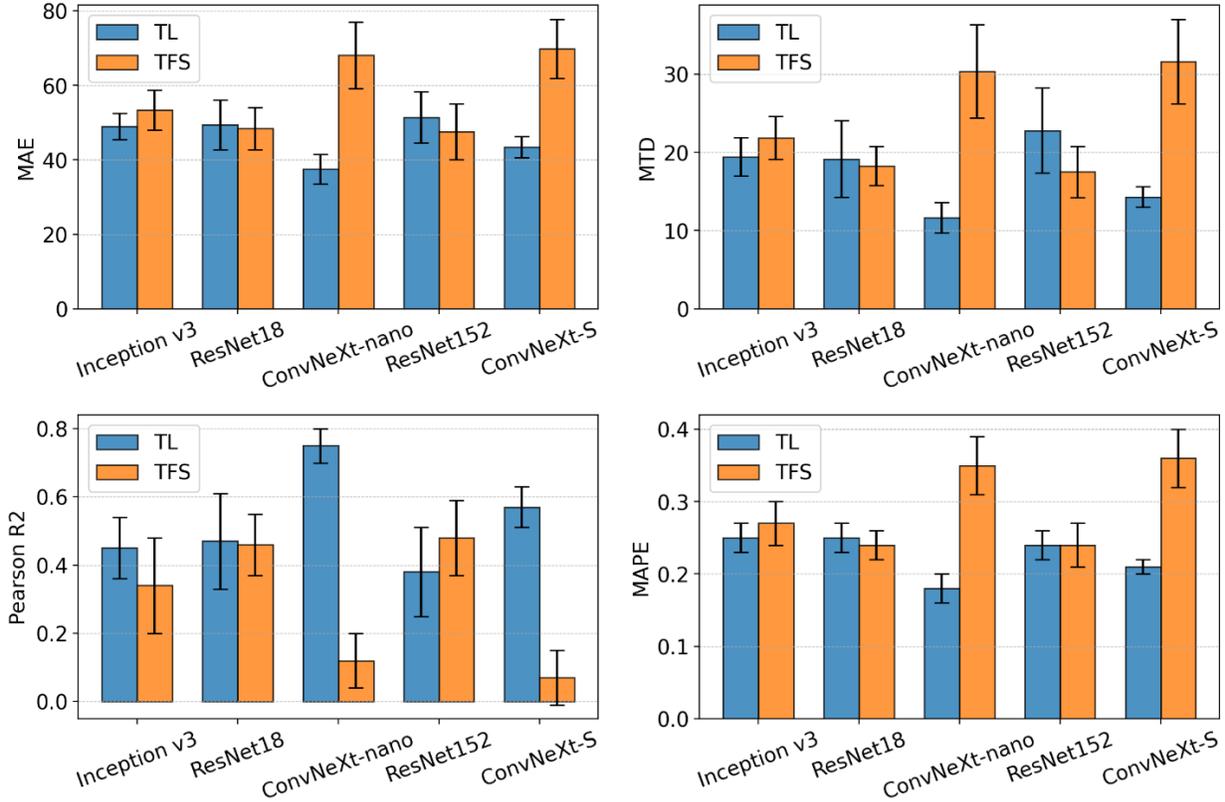

**Fig. 4. Comparative evaluation of error metrics across TL and TFS models on the test dataset**

Figure 4 shows that, in general, TL models surpass TFS models when evaluated by the selected error measures. This result highlights the efficacy of TL in applying pre-trained knowledge to enhance CNN generalization and performance for the considered dataset. Among the TL models, ConvNeXt-nano (TL) exhibited optimal performance, securing the lowest MAE (37.51 ± 4.02) and MTD (11.65 ± 1.94), MAPE (0.18 ± 0.02), and the highest $R^2$ (0.75 ± 0.05). Inception v3 (TL) and ResNet18 (TL) also performed well with comparable error metrics. When analyzing the TFS models, ResNet18 (TFS) provides a higher performance, although not reaching the ConvNeXt-nano (TL) results. Once more, this result indicates that CNN architectures that were fine-tuned with TL are more effective for this study. A possible reason for the lower performance of ResNet18 (TFS) can be linked to the random weight initialization, potentially resulting in slower convergence leading to a suboptimal performance. ConvNeXt-S and ConvNeXt-nano showed the weakest performances among the TFS models, resulting in the highest MAE, MTD, and MAPE and the lowest $R^2$. Again, this result can be related to the random weight initialization and the absence of prior knowledge. The ConvNeXt-nano model was the most efficient CNN architecture for SVI prediction when fine-tuned with TL.

**3.2 Predictive performance of the selected CNN**

The ConvNeXt-nano CNN model underwent initial analysis for its capability to predict SVI during training iterations, adhering to the previously outlined configuration. Figure 5 depicts the scatter plot of the SVI predictions by the ConvNeXt-nano CNN during the cross-validation phase.

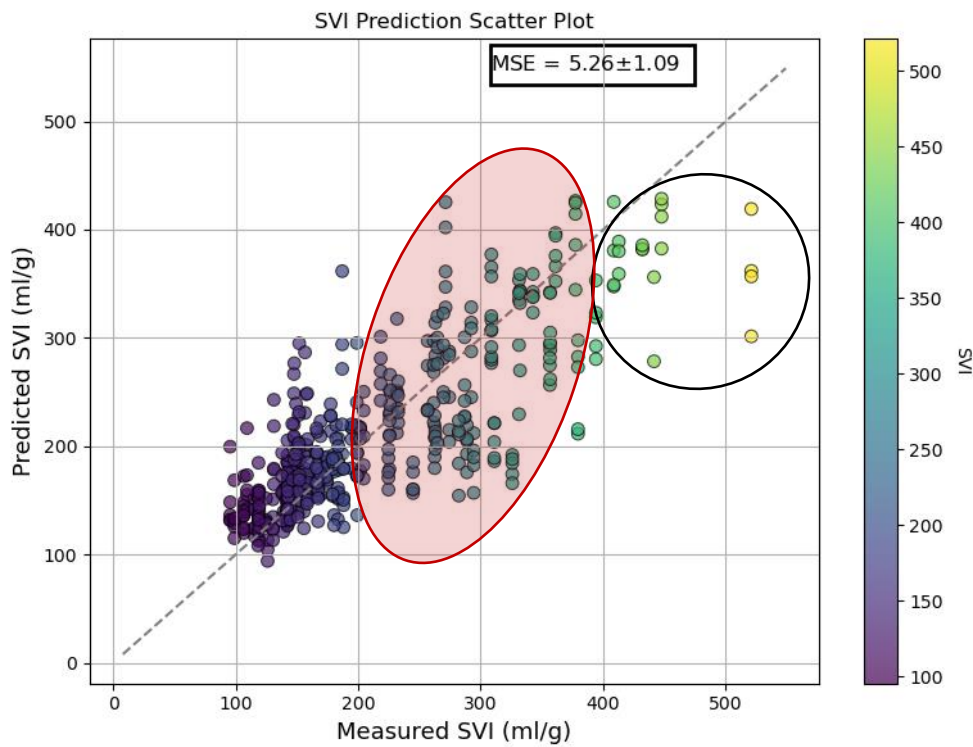

**Fig. 5. Ground-Truth vs. Predicted SVI Values using ConvNeXt-nano CNN during cross-validation**

An important observation from Fig. 5 is that although the model performs satisfactorily on samples with lower SVI levels, it struggles with samples with SVI values above a certain threshold (400 ml/g), marked by a black circle. In these cases, actual SVI values often exceed the model's predictions. This issue may be linked to a bias in the training data, which mainly includes instances of lower SVI values, limiting the ConvNeXt-nano model's capability to analyze higher SVI levels effectively. Another intriguing observation is the variable prediction accuracy when the ground-truth SVI values are consistent, but the predictions differ, which is evident among observations in the red-coloured oval in Fig. 5. Despite identical SVI measurements, variations in morphological features, floc and filament structures across images can result in different SVI predictions. It can be due to the intrinsic multi-modality in the SVI values, where one SVI value can reflect multiple morphological features. While SVI is a standard metric for assessing the settling characteristics of activated sludge, it may not fully capture all characteristics seen in microscopic evaluations. Therefore, although this study uses SVI as the primary prediction variable, the results suggest that microscopy images might offer a more thorough and detailed analysis of sludge morphology than SVI. Extending the research to apply the CNN-based approach to different or multiple settling indices might reveal more details about settling characteristics.

## 3.3 Performance of the selected CNN

The performance of ConvNeXt-nano was put to a real-world test in predicting SVI values during FB events. Figure 6 displays the ConvNeXt-nano CNN model predictions and SVI measurements. A red line represents the average prediction for each day, while the shaded region indicates the standard deviation since multiple images were available per day. An excellent model predictive performance can be observed for lower SVI values during non-bulking periods, while the model could provide valuable insights into the early stages of elevating SVI values, which can serve as preliminary signals of FB events. In the context of WWTP operations, a sudden or gradual increase in SVI values can serve as an early warning sign for FB events. The model predictions show a noticeable increase in SVI values in some incidents. This emphasizes the model's capability to be used as an early warning system. It is crucial to clarify that while the model acts as a critical operational tool by detecting elevating trends in SVI before FB, allowing operators to take pre-emptive actions, it is not designed to forecast specific SVI values without corresponding future image data. Therefore, the model's application dwells in its capacity to detect initial SVI elevations to instigate timely interventions rather than predicting exact future SVI values or FB events.

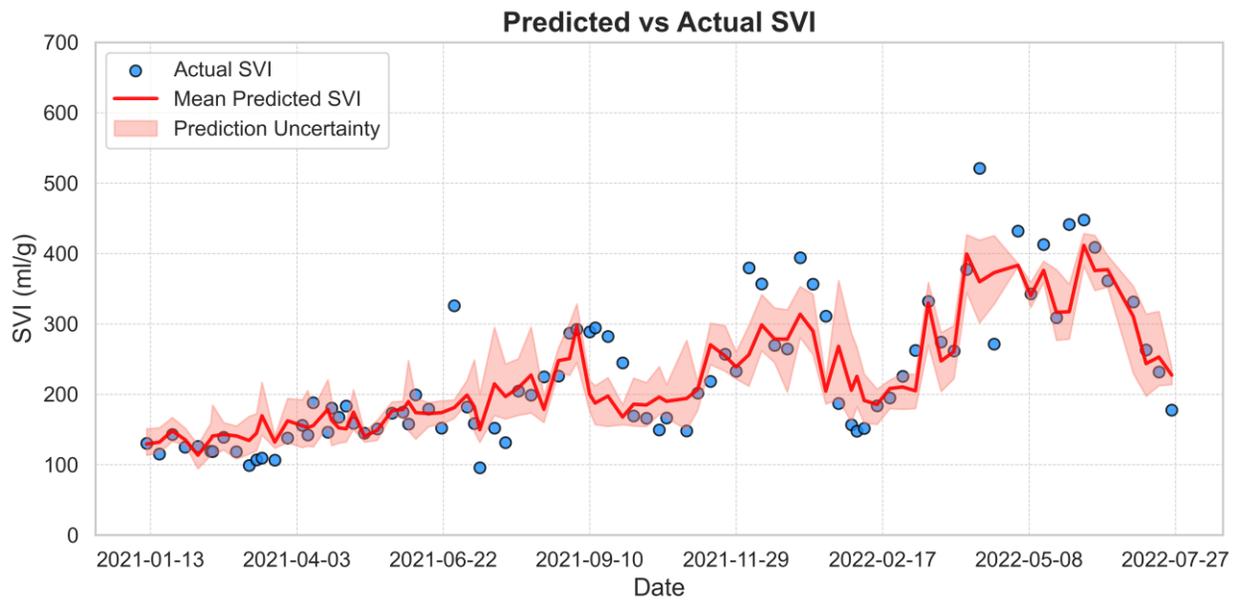

**Fig. 6. Plot analysis of ConvNeXt-nano predictions for SVI**

## 4. Conclusions

Prompt detection and forecasting of activated sludge settling issues, including filamentous bulking (FB), can prevent operational challenges and financial implications. In this study, we introduced an innovative method for assessing activated sludge settling characteristics using computer vision techniques. Transfer learning of deep convolutional neural network (CNN) models was proposed to study the morphological properties of flocs and filaments in microscopy images. The offline microscopy images dataset was collected over two years, with weekly sampling at a full-scale industrial WWTP in Belgium. We explored different data augmentation methods to make the model more adaptable to filament and floc structure variations, enhancing its versatility. A comparative analysis of various CNN architectures, including Inception v3, ResNet18, ResNet152, ConvNeXt-nano, and ConvNeXt-S, was conducted to evaluate their performance in predicting sludge volume index (SVI). The findings showed that transfer learning significantly reduced the model's training time, reaching optimal results in just 30 epochs. This made the model both computationally efficient and practically applicable. Notably, the ConvNeXt-nano model outperformed other tested models across critical metrics, including MAE (37.51 ± 4.02), MTD (11.65 ± 1.94), MAPE (0.18 ± 0.02), and $R^2$ (0.75 ± 0.05). A real-world application was demonstrated during filamentous bulking (FB) events and normal conditions recorded during the sampling campaigns. The model could effectively signal early signs of FB indicated by surges in SVI predictions. This makes the ConvNeXt-nano a potent tool for proactive response planning in WWTPs. The model's performance on a limited dataset suggests its broader applicability.

# 6. Acknowledgments

L. Scabini acknowledges funding from the São Paulo Research Foundation (FAPESP), Brazil (Grants #2019 /07811-0, #2021/09163-6, and #2023/10442-2).

B. De Baets received funding from the Flemish Government under the "Onderzoeksprogramma Artificiële Intelligentie (AI) Vlaanderen" programme.

9. Cenens, C., Van Beurden, K.P., Jenné, R., Van Impe, J.F., 2002. On the development of a novel image analysis technique to distinguish between flocs and filaments in activated sludge images. Water Sci. Technol. 46, 381–387.
10. Chai, J., Zeng, H., Li, A., Ngai, E.W., 2021. Deep learning in computer vision: A critical review of emerging techniques and application scenarios. Mach. Learn. Appl. 6, 100134.
11. Chan, J.Y.-L., Bea, K.T., Leow, S.M.H., Phoong, S.W., Cheng, W.K., 2023. State of the art: a review of sentiment analysis based on sequential transfer learning. Artif. Intell. Rev. 56, 749–780. https://doi.org/10.1007/s10462-022-10183-8
12. Comas, J., Rodríguez-Roda, I., Gernaey, K.V., Rosen, C., Jeppsson, U., Poch, M., 2008. Risk assessment modelling of microbiology-related solids separation problems in activated sludge systems. Environ. Model. Softw. 23, 1250–1261.
13. Costa, J.C., Mesquita, D.P., Amaral, A.L., Alves, M.M., Ferreira, E.C., 2013. Quantitative image analysis for the characterization of microbial aggregates in biological wastewater treatment: a review. Environ. Sci. Pollut. Res. 20, 5887–5912.
14. Costa, J.G., Paulo, A.M., Amorim, C.L., Amaral, A.L., Castro, P.M., Ferreira, E.C., Mesquita, D.P., 2022. Quantitative image analysis as a robust tool to assess effluent quality from an aerobic granular sludge system treating industrial wastewater. Chemosphere 291, 132773.
15. Da Motta, M., Pons, M.-N., Roche, N., Vivier, H., 2001. Characterisation of activated sludge by automated image analysis. Biochem. Eng. J. 9, 165–173.
16. Deng, J., Dong, W., Socher, R., Li, L.-J., Li, K., Fei-Fei, L., 2009. Imagenet: A large-scale hierarchical image database, in: 2009 IEEE Conference on Computer Vision and Pattern Recognition. Ieee, pp. 248–255.
17. Dias, P.A., Dunkel, T., Fajado, D.A., Gallegos, E. de L., Denecke, M., Wiedemann, P., Schneider, F.K., Suhr, H., 2016. Image processing for identification and quantification of filamentous bacteria in in situ acquired images. Biomed. Eng. OnLine 15, 1–19.
18. Dosovitskiy, A., Beyer, L., Kolesnikov, A., Weissenborn, D., Zhai, X., Unterthiner, T., Dehghani, M., Minderer, M., Heigold, G., Gelly, S., Uszkoreit, J., Houlsby, N., 2021. An Image is Worth 16x16 Words: Transformers for Image Recognition at Scale.
19. Dung, C.V., 2019. Autonomous concrete crack detection using deep fully convolutional neural network. Autom. Constr. 99, 52–58.